\documentclass{article}

\PassOptionsToPackage{numbers, compress}{natbib}

      \usepackage[final]{neurips_2021}

\usepackage[utf8]{inputenc} 
\usepackage[T1]{fontenc}    
\usepackage{url}            
\usepackage{booktabs}       
\usepackage{amsfonts}       
\usepackage{nicefrac}       
\usepackage{microtype}      
\usepackage{xcolor}         
\usepackage{amsmath}
\usepackage{bm}
\usepackage{multirow}
\usepackage[pagebackref=true,breaklinks=true,colorlinks,bookmarks=false]{hyperref}
\usepackage[pdftex]{graphicx}

\usepackage{wrapfig}

\usepackage{multicol}
\title{Is Object Detection Necessary for Human-Object Interaction Recognition?}

\author{  
  Ying Jin\textsuperscript{\dag\ddag} \ Yinpeng Chen\textsuperscript{\dag} \ Lijuan Wang\textsuperscript{\dag} \ Jianfeng Wang\textsuperscript{\dag} \\ \textbf{Pei Yu\textsuperscript{\dag} \ Zicheng Liu\textsuperscript{\dag} \ Jenq-Neng Hwang\textsuperscript{\ddag}} \\

  \textsuperscript{\dag}Microsoft \ \ \ \ 
  \textsuperscript{\ddag}University of Washington\\

  \texttt{\{ying.jin,yiche,lijuanw,jianfw,peyu,zliu\}@microsoft.com}\\
  \texttt{\{jinying,hwang\}@uw.edu}
  \vspace{-1em}
}

\begin{document}
\maketitle
\begin{abstract}
  This paper revisits human-object interaction (HOI) recognition at image level \textit{without} using supervisions of object location and human pose. We name it detection-free HOI recognition, in contrast to the existing detection-supervised approaches which rely on object and keypoint detections to achieve state of the art. With our method, not only the detection supervision is evitable, but superior performance can be achieved by properly using image-text pre-training (such as CLIP) and the proposed \textit{Log-Sum-Exp Sign (LSE-Sign)} loss function. Specifically, using text embeddings of class labels to initialize the linear classifier is essential for leveraging the CLIP pre-trained image encoder. In addition, LSE-Sign loss facilitates learning from multiple labels on an imbalanced dataset by normalizing gradients over all classes in a softmax format. Surprisingly, our detection-free solution achieves 60.5 mAP on the HICO dataset, outperforming the detection-supervised state of the art by 13.4 mAP. 
\end{abstract}

\section{Introduction}

\begin{wrapfigure}[21]{r}{0.5\textwidth}
  \vspace{-1.5em}
  \centering
  \includegraphics[width=0.715\linewidth]{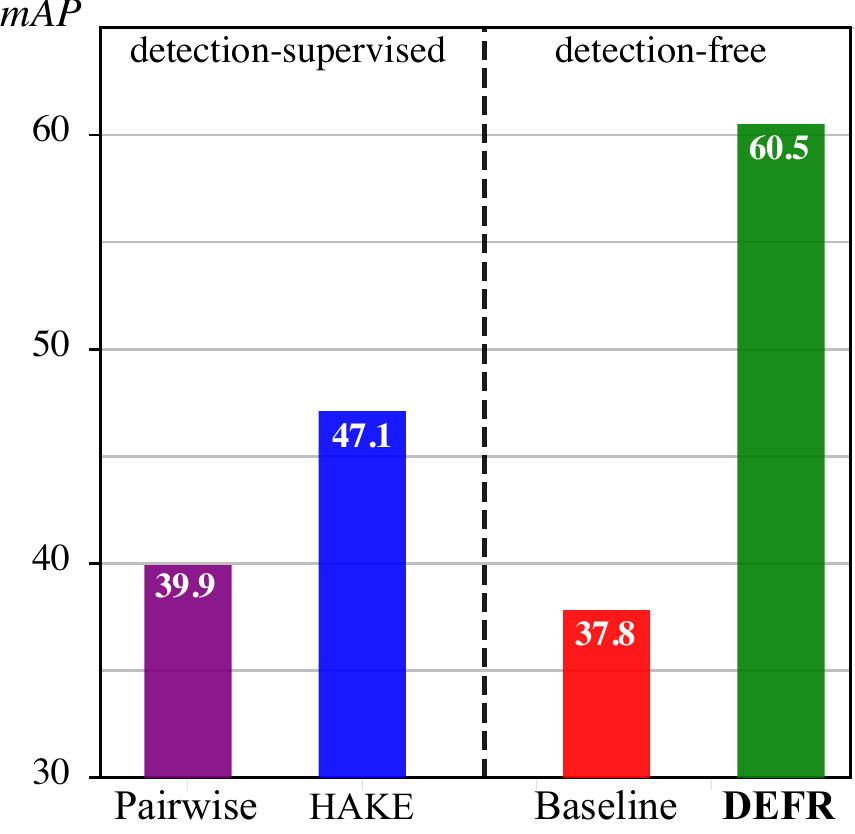}

  \caption{\textbf{\small Detection-Supervised v.s Detection-Free} HOI recognition. Although the detection-free baseline (using ImageNet pre-trained ViT-B/32 \cite{vit20} as backbone and binary cross entropy loss) degrades severely from the detection-supervised HAKE \cite{hake19}, DEFR, our method, achieves significant improvement by leveraging image-text pre-training and the proposed LSE-Signed loss.}
  \label{fig:1}
\end{wrapfigure}

Human-Object Interaction (HOI) recognition has drawn significant research interest in recent years as it plays an important role in human-centric scene understanding. It aims to retrieve \textit{<verb, object>} pairs for a given image. Existing studies \cite{pastanet20,pairwise18,rcnn15, hico15, sgg-imp17, sgg-knowledge-embed19, sgg-commonsense20} mostly rely on object detectors to first detect humans and objects, and then infer interactions at the instance level. In contrast, this paper aims at a \textit{detection-free} solution without affecting the prediction accuracy.

By eliminating the supervision of object and human keypoint detection, detection-free HOI recognition significantly simplifies the pipeline. However, the problem is very challenging as the positions of humans and objects are unknown. A naive solution with a vision transformer backbone pre-trained on ImageNet-1K experiences a considerable performance degradation compared to the state of the art \cite{hake19} (37.8 mAP \textit{vs.} 47.1 mAP in \autoref{fig:1}).

In this paper, we show that the performance of detection-free approach can be significantly boosted by using \textit{image-text pre-training} and the \textit{log-sum-exp sign} loss function. The former not only facilitates learning of relationships between humans and objects, but also encodes the structure of HOI classes. The latter successfully handles multiple HOI labels per image. Our method, named DEtection FRee (DEFR) HOI recognition, is based upon these two pillars.

\begin{figure}[t]
  \centering
  \includegraphics[width=0.8\linewidth]{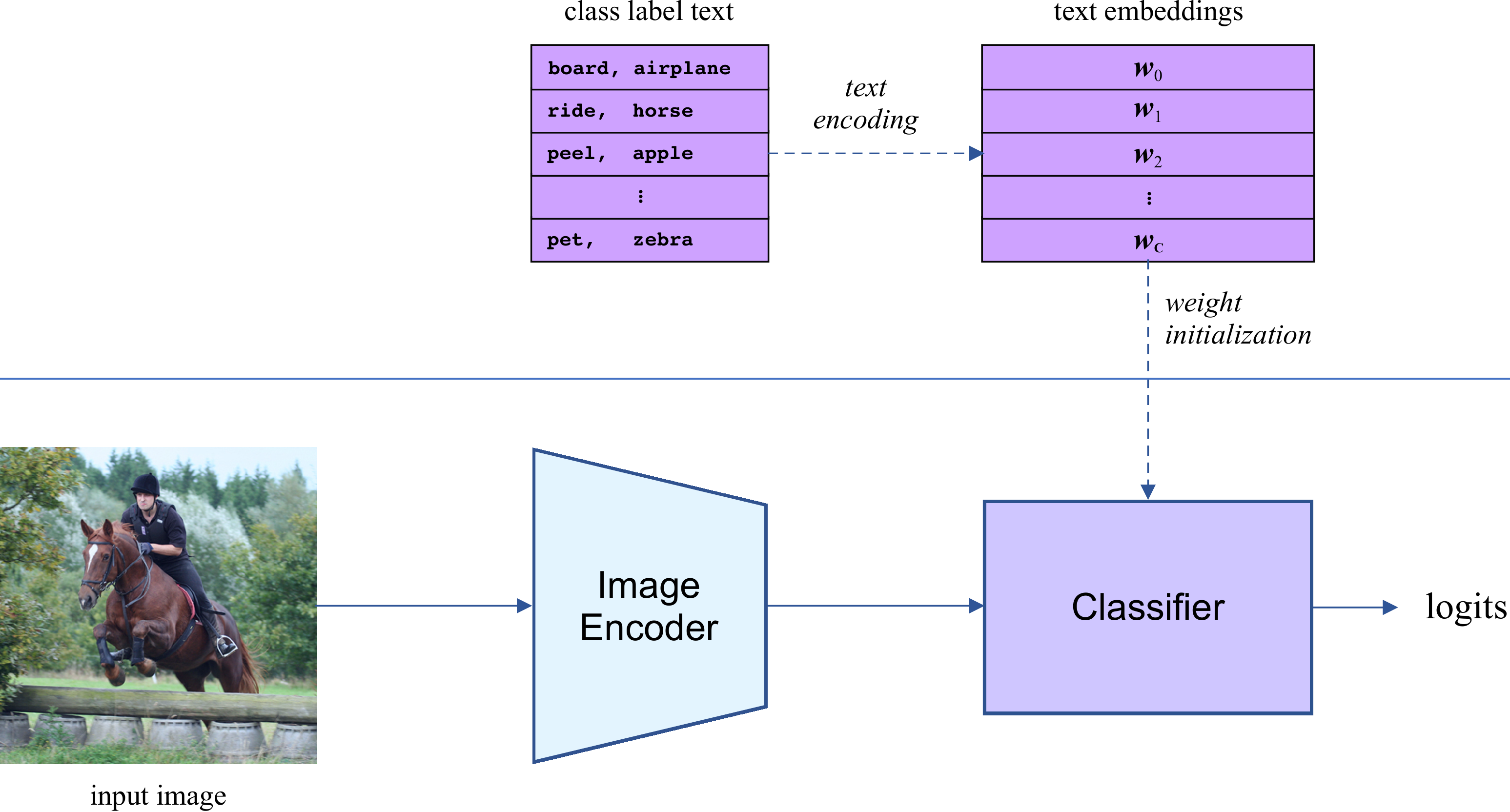}
  \caption{\textbf{DEtection FRee (DEFR) HOI recognition pipeline}. It has an image encoder and a linear classifier. The weight of the linear classifier is initialized by $w$, the text embeddings of class labels encoded by BERT or CLIP's text encoder. We call this \textit{embedding initialization}, detailed in \autoref{sec:embedding_init}. Compared to the detection-supervised approaches, DEFR significantly simplifies the pipeline.}
  \label{fig:pipeline}
\end{figure}

Regarding image-text pre-training, a significant performance boost is achieved when \textit{embedding initialization} (see \autoref{fig:pipeline}) is jointly used with CLIP \cite{clip21} pre-trained image backbone. Furthermore, embedding initialization works with image-alone pre-trained backbones (like ImageNet pre-trained models), contributing a gain of 10+ mAP. Our use of the text encoder (in embedding initialization) differs from the original use (for zero-shot classification) in CLIP, and we show that this is crucial for HOI recognition in the fully supervised setup.

In terms of loss function, we propose Log-Sum-Exp Sign (LSE-Sign) loss to handle multiple labels per image. Compared to binary entropy loss which considers each class independently, LSE-Sign loss enables interaction between classes by normalizing the gradients over all classes in a softmax format. The softmax format encourages more attention towards the class with maximum loss.

Based upon these two new findings, our detection-free solution (DEFR) outperforms previous detection-supervised approaches \cite{pairwise18,pastanet20,hake19} by a clear margin with either image-text (CLIP) or image-alone (ImageNet-1K/21K) pre-trained backbones. Our best model achieves 60.5 mAP on the HICO \cite{hico15} dataset, surpassing the state of the art \cite{hake19} that uses both object and human keypoint detection, by 13.4 mAP.

\section{Related work}

\textbf{HOI Recognition:}
For HOI recognition, a key challenges is the co-occurrence of multi-human and multi-object when the positions of HOIs in an image are not available. Existing work \cite{rcnn15,transfer16,pairwise18} depend on off-the-shelf object detector or keypoint detector and usually requires Multiple Instance Learning \cite{mil98}. \citet{attentionalpooling17} predict attention maps per HOI and require no object detection or MIL. PastaNet \cite{pairwise18} and HAKE \cite{hake19} achieve state-of-the-art performance by conducting body part-level action classification first, and additional training data is required.

\textbf{HOI Detection:}
Instance level HOI detection on HICO-DET \cite{hicodet18} is attracting growing interest. Existing work can be categorised into three streams. Two-stage methods \cite{ican18,kaiming_hoi18,drg20,fcm20,pastanet20,scg20} require object detection prior to HOI classification to extract regional features. One-stage methods \cite{liao2020ppdm,uniondet20,ggnet21} execute object detection and HOI detection concurrently and pair them afterwards. Recent studies \cite{asnet21,hotr21,e2e21,qpic21} achieve end-to-end HOI detection with a DETR \cite{detr2020} style network and benefit from the wider perception field of transformers \cite{qpic21}.

\textbf{Scene Graph Generation:}
As in connection with HOI, scene graphs \cite{vg17} are also pairwise relationships. A scene graph is a graph representation in which the nodes are objects and edges are relationships. The IMP \cite{sgg-imp17} method detects objects first and region features are then iteratively fine-tuned in a graphical network. To reduce the compute, \citet{sgg-factorizable18} constructs sparsely connected sub-graphs, and \citet{sgg-grcnn18} applies a Relationship Proposal Network. \citet{sgg-neuralmotif18} reveals the strong statistical bias in the Visual Genome \cite{vg17} dataset, and \citet{sgg-unbiased20} suggests unbiased post-processing to neutralize the model from biased training. \citet{sgg-knowledge-embed19} and \citet{sgg-commonsense20} use frequency priors from the dataset and text as external knowledge or common sense to improve performance.

\textbf{Vision-Language Pre-training:}
Although leveraging natural language in HOI and scene graph tasks is not new \cite{sgg-knowledge-embed19, sgg-commonsense20,pastanet20}, these approaches are limited to the use of frequency priors and word embeddings and the gain is quite limited. In image-text pre-training, images and text are mapped to the same vector space, and matching pairs have higher cosine similarities. Vision-Language pre-training has been proved useful in image captioning \cite{oscar20,vivo20}. CLIP from OpenAI \cite{clip21} are publicly available and are trained on 400 million (text, image) pairs. CLIP contains an image encoder and a text encoder.

\section{DEFR: DEtection-FRee human-object interaction recognition}

In this section, we introduce our detection-free solution (DEFR) for human-object interaction (HOI) recognition. The goal of DEFR is to eliminate the supervision of object and human keypoint detection without performance degradation. Surprisingly, we find that this can be achieved through the use of image-text pre-training and the proposed log-sum-exp sign (LSE-Sign) loss function.

\subsection{Detection-free HOI recognition pipeline}
Our DEFR method has a simple pipeline (shown in \autoref{fig:pipeline}), including a vision transformer \cite{vit20} (ViT-B/32) as the backbone and a linear layer as the classier. The classifier is text embedding initialized, instead of the current go-to approach of randomly initialized \cite{xavier10,kaiminginit15}. Compared to the detection-supervised methods \cite{pairwise18,drg20,pastanet20,hake19}, our method simplifies the pipeline significantly and enables learning in an end-to-end manner.

Despite the simple pipeline, the problem becomes challenging due to the elimination of detection supervision, which provides accurate spatial information about humans and objects. To validate this, we fine-tune a vision transformer (ViT-B/32) backbone pre-trained on ImageNet-1K and a randomly initialized linear classifier by using binary cross entropy loss on the HICO dataset. A severe degradation (to 37.8 mAP) is observed when compared with the detection-supervised approaches (47.1 mAP in HAKE \cite{hake19}).

\subsection{Image-text pre-training}
\label{sec:embedding_init}

\begin{figure}
  \setlength{\tabcolsep}{1pt}
  \centering
  \caption{\textbf{Visualization of CLS attention map}. Three images in a group are input image, attention map of the pre-trained CLIP, and attention map of DEFR (fine-tuned). After fine-tuning, the attention activates more on the HOI related objects and activates less elsewhere.}
  \vspace*{1em}
  \includegraphics[width=0.8\linewidth]{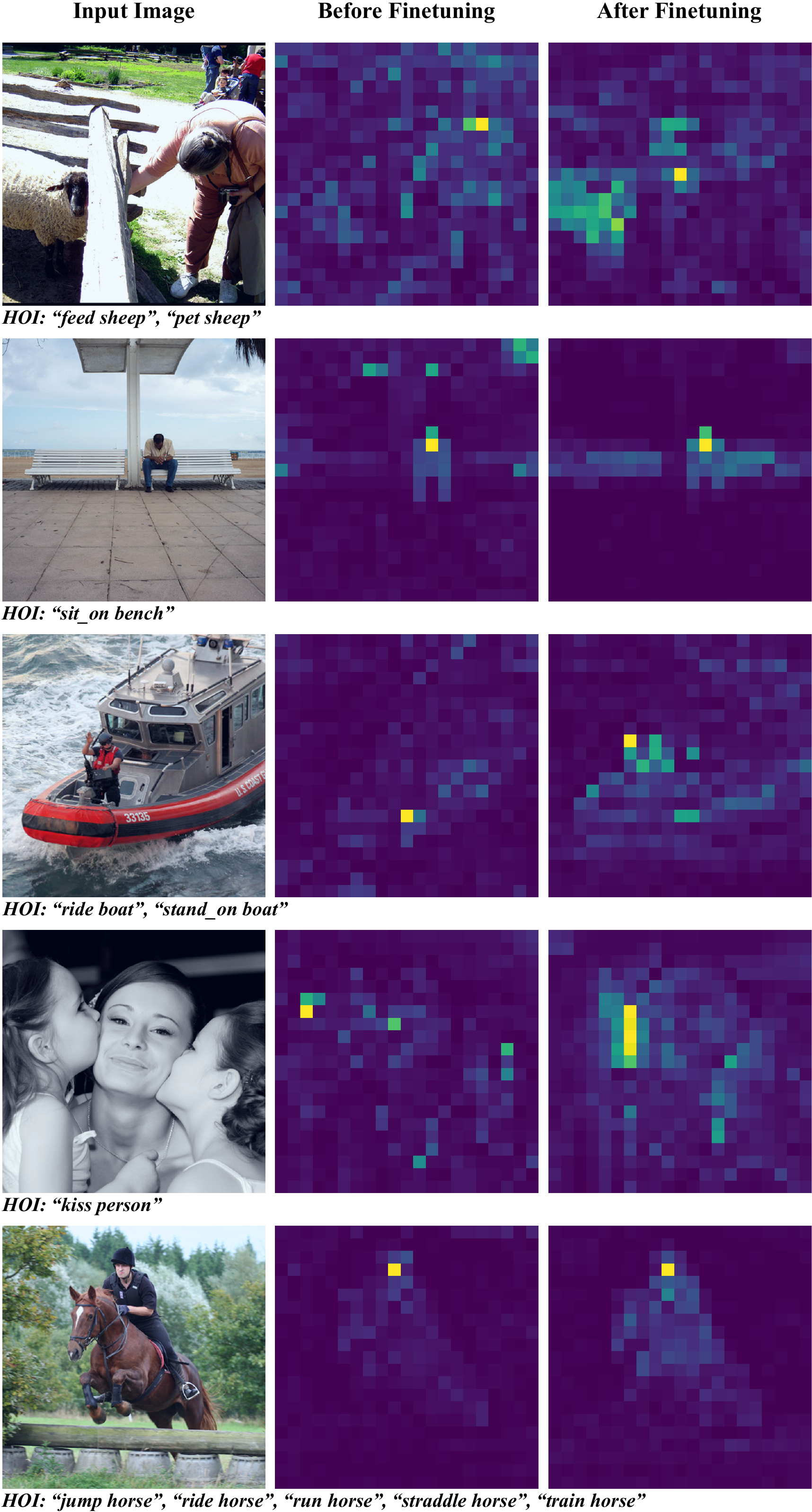}
  \label{fig:vis-cls-attn}
\end{figure}

Image-text pre-trained ViT provides a natural compensation for the elimination of detection supervision, since it implicitly relates features of humans and objects through self-attention mechanism. To demonstrate this, we apply a CLIP \cite{clip21} pre-trained model on the task of HICO dataset. As shown in \autoref{fig:vis-cls-attn}, people and objects get more attention from the class label than the background.

\begin{figure}[hb]
  \centering
  \includegraphics[width=1\linewidth]{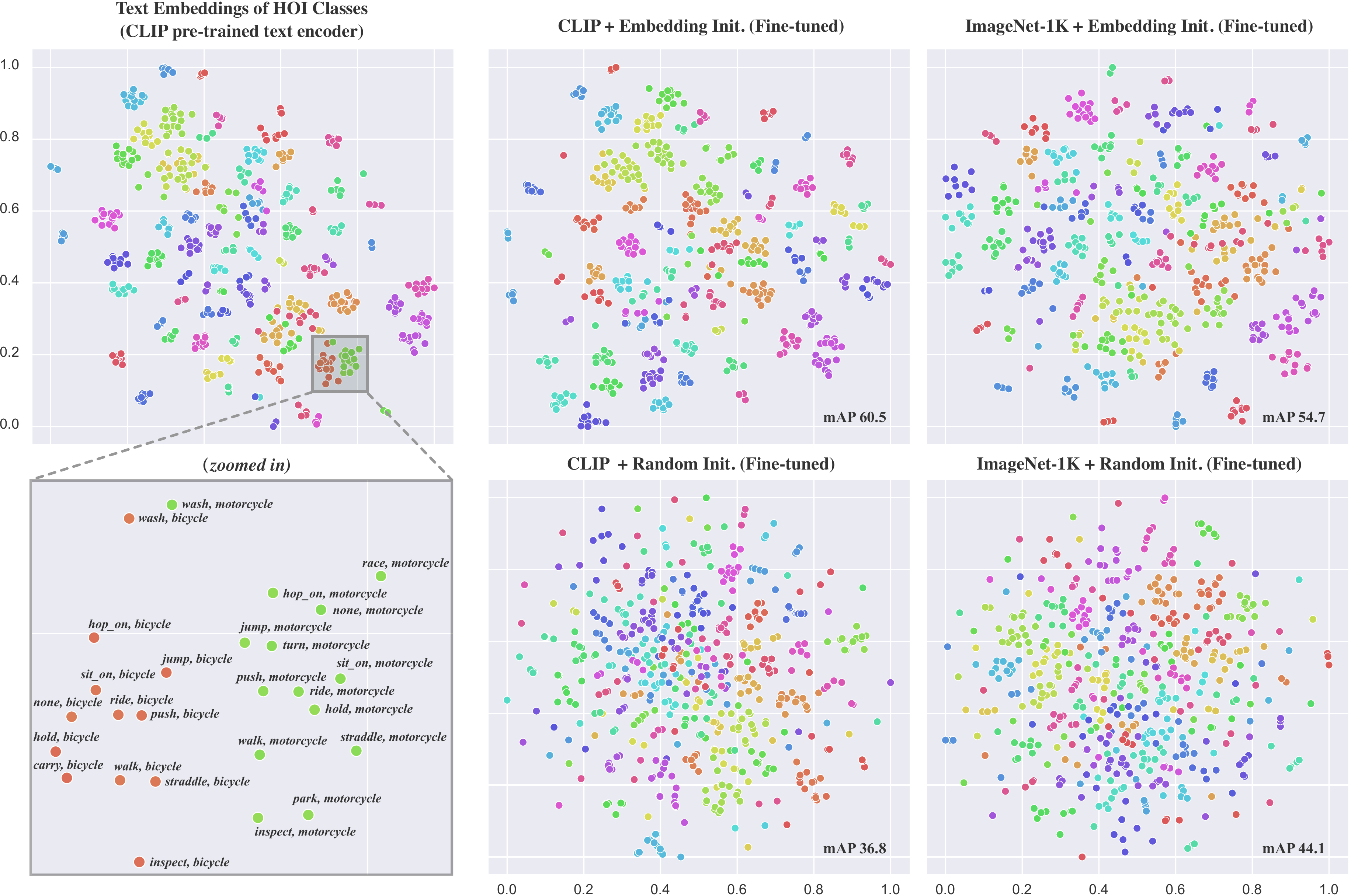}
  \vspace{1mm}
  \caption{\textbf{TSNE visualization of classifier weights.} \textbf{Left Column:} the CLIP-encoded text embeddings of 600 HOI class labels. Each point represents the embedding of an HOI class. Text embeddings encode the structure of HOI classes: as is shown in the zoomed view, points that represent similar semantic meanings locate closely. \textbf{Middle Column:} the weight vectors of 600 HOI classes (i.e. row vectors in linear classifier's weight matrix) after fine-tuning, when CLIP pre-trained ViT-B/32 backbone is used. The top and bottom use embedding initialization and random initialization for the linear classifier, respectively. \textbf{Right Column:} the weight vectors of 600 HOI classes after fine-tuning, when ImageNet-1K pre-trained ViT-B/32 backbone is used. The top and bottom use text embedding and random initialization for the linear classifier, respectively. Text embeddings are clearly clustered before fine-tuning, and the overall structure changes slightly after fine-tuning. However, this clustered structure is not clearly learned when fine-tuned with random initialization, reflected in lower model performance (see mAP results in the plots).}
  \label{fig:tsne-class-embedding}
\end{figure}

\textit{{Embedding initialization}} complements image-text pre-trained backbones. Embedding initialization is to initialize the linear classifier with text embeddings of class labels. Specifically, we convert the HOI classes (e.g. "ride bicycle") to sentences (e.g. "a person riding a bicycle") and generate their text embeddings with a text encoder. Then text embeddings are normalized and used as the initial weight in the linear layer. The rationale can be explained from a proxy learning perspective. Assume bias is not used, the output logit for the $i^{th}$ class is the dot product of the image feature and $w_i$, a row vector in the classifier's weight. Since dot product is the unnormalized cosine similarity, $w_i$ is also considered as the proxy for a given class \cite{circleloss20,normface17}. For a multi-label task like HOI recognition, 
the image feature need to move toward multiple related proxies in their vector space during fine-tuning. This could be hard when proxies are randomly scattered in the vector space. Embedding initialization provides an initial spatial structure for the proxies with their semantic distance into account (see \autoref{fig:tsne-class-embedding}), enabling the model to leverage both vision and language modalities.

We find that embedding initialization fires a performance boost of 10+ mAP. More importantly, the applicability generalizes to image-alone pre-trained backbones (e.g. ImageNet pre-trained ViT-B/32), and the gain is even more significant when the backbone and the text encoder that generates the embeddings are jointly trained (e.g. CLIP). We believe this is because the CLIP text encoder learns a cross-modality aligned representation, which benefits HOI recognition. \autoref{fig:tsne-class-embedding} visualizes the class embedding before (left-column) and after fine-tuning with both CLIP (middle-column) and ImageNet-1K (right-column) pre-trained backbones. HOI class embeddings are clearly clustered before fine-tuning, and the overall structure changes slightly after fine-tuning. Interestingly, the structure holds well even for the backbone pre-trained on images alone (ImageNet-1K). However, this structure is difficult to learn when fine-tuning with random initialization, reflected in lower performance (shown in \autoref{fig:tsne-class-embedding} bottom row).

This finding demonstrates another application of text embeddings in linear classifier, which is complementary to its original usage in \cite{clip21} for zero-shot classification. Please note that the technical contribution of this paper is \textit{not} image-text pre-training, but rather unlocking its power for detection-free HOI recognition.

\subsection{Log-Sum-Exp Sign (LSE-Sign) Loss}
We propose a log-sum-exp sign (LSE-Sign) loss function to facilitate learning from multiple labels, as an input image can have multiple interactions (e.g. \textit{cut carrot, hold carrot, peel carrot}). Let $\bm{x}$ denote a feature vector from the backbone and $\bm{y}=\{y_1, y_2,\dots,y_C\}$ denote multiple class labels, where $C$ is the number of HOI classes. $y_i \in \{1, -1\}$, indicating positive and negative classes, respectively. For CLIP backbones, we follow \cite{clip21} to normalize the output per class and re-scale it as follows:

\begin{equation}
  \label{eq:si}
  s_i = \gamma\frac{\bm{x}^T \bm{w}_i}{\Vert \bm{x} \Vert \Vert \bm{w}_i \Vert},
\end{equation}

where $\bm{w}_i$ is the $i^{th}$ row in the weight matric of the linear classifier, corresponding to the $i^{th}$ HOI class. $\gamma$ is a scalar hyper-parameter, controlling the output range. It is equivalent to scale cosine similarity such that the output value is between $\pm\gamma$. The losses of both positive and negative classes can be united into $e^{-y_is_i}$, where the label $y_i$ controls the sign. The overall loss is defined as the log-sum-exp function of $-y_is_i$ as follows:

\begin{equation}
  \label{eq-loss}
  \mathcal{L} = \log\left(1+\sum_{i=1}^C e^{-y_is_i}  \right) = \log\left(1+\sum_{i\in pos} e^{-s_i} + \sum_{i\in neg} e^{s_i} \right),
\end{equation}

where the first term 1 in the log function sets the zero lower bound for the loss. Log-sum-exp is a smooth approximation of the maximum function, and its gradient is the softmax function as follows:

\begin{equation}
  \label{eq-gradient}
  \frac{\partial \mathcal{L}}{\partial s_i} = \frac{-y_ie^{-y_is_i}}{1+\sum_{j=1}^C e^{-y_js_j}} =
  \frac{-1_{i \in pos}e^{-s_i}+1_{i \in neg}e^{s_i}}{1+\sum_{i\in pos} e^{-s_i} + \sum_{i\in neg} e^{s_i}}, 
\end{equation}
where $1_{i \in pos}$ is a Dirac delta function that returns $1$ if the $i^{th}$ class is positive and $0$ otherwise. Compared to the binary cross entropy loss that considers each class separately, the LSE-Sign loss considers the dependency across classes as the magnitude of the gradients over all classes are normalized and distributed by a softmax function. This facilitates learning from multiple labels as it encourages learning for classes with larger loss values and suppresses the learning for classes with smaller loss values. This feature is especially helpful on imbalanced datasets, where few shot classes need amplified gradients during back-propagation to properly train.

\section{Experiments}
\textbf{Dataset:}
We conduct experiments on the HICO \cite{hico15} dataset to evaluate our DEFR method on image-based HOI recognition. HICO contains 600 HOI categories which have 117 unique verbs and 80 COCO \cite{coco14} object classes. Each image has one or multiple HOI classes. The training set contains $38,116$ images and test set contains $9,658$ images. We randomly reserve 10\% images from the training set for validation.

\textbf{Pre-training:} The backbone is ViT-B/32, and we investigate three different strategies to pre-train the backbone at resolution 224.
\begin{enumerate}
  \item Image classification task on ImageNet-1K or ImageNet-21k referred to as CLS1K and CLS21K, respectively.
  \item Masked language modeling task motivated by~\cite{kim2021vilt}, where the network input includes both an image and the associated text. Both modalities are fully attended with each other in each transformer block. The pre-training datasets are Google Conceptual Captions~\cite{SoricutDSG18}, SBU~\cite{OrdonezKB11}, COCO~\cite{coco14} and Visual Genome~\cite{vg17}. This is referred to as MLM.
  \item Image–text contrastive learning task based on CLIP~\cite{clip21}. The image encoder is jointly trained with a text encoder and the two modalities are contrasted on the encoder output representations. Only the image encoder is used as the backbone. Here, we directly use the released CLIP model \footnote{https://github.com/openai/CLIP}, and reuse the term CLIP to refer to the image encoder for initialization.
\end{enumerate}

\textbf{Fine-tuning:} All models have a ViT-B/32 backbone fine-tuned at resolution 672 with the Adam \cite{adam14} optimizer with no weight decay. We use a batch size of 128 on 8 V100 GPUs. We set the learning rate as 1e-5 when using the CLIP backbone and 1e-4 otherwise, and use cosine scheduling with warm restarts every 5 epochs. The best model is fine-tuned for 10 epochs. Data augmentations of random color jitter, horizontal flip and resized crop are used. To reduce class imbalance, we apply over-sampling so that each class has at least 40 training images in each epoch. For embedding initialization, similar to \cite{clip21}, we convert HOI class labels (e.g. "ride bicycle") to prompts (e.g. "a person riding a bicycle") then generate text embeddings with BERT \cite{bert18} or CLIP's text encoder.

\subsection{Comparison with State of the Art}
\autoref{tab:mAP} compares our detection-free method (DEFR) with the prior works that use supervision of either object detection or human keypoint detection. Our DEFR method achieves significantly better accuracy without detection supervision. Specifically, DEFR achieves 60.5 mAP, gaining over the state-of-the-art HAKE \cite{hake19} by 13.4 mAP.

\begin{table}[t]
  \centering
  \caption{\textbf{Comparison with the state of the art on HICO}. Supervisions include \emph{Bbox}: object detection, \emph{Pose}: human keypoints, \emph{PaSta} \cite{pastanet20}: additional training data of part level actions, e.g. \textit{<hand, hold, something>}. Compared to the detection-supervised approaches, our DEFR method frees the supervision of detection while achieves a significant boost on accuracy.
  }
  \label{tab:mAP}

  \begin{tabular}{ l c c c c }\toprule
    \multirow{2}{*}{Method}                             & \multicolumn{3}{c }{\textbf{Supervision}} & \multirow{2}{*}{mAP}                                     \\
    \cmidrule(lr){2-4}
                                                        & \textbf{Bbox}                             & \textbf{Pose}        & \textbf{PaSta} &                 \\
    \cmidrule(lr){1-1}
    \cmidrule(lr){2-4}
    \cmidrule(ll){5-5}
    R*CNN \cite{rcnn15}                                 & \checkmark                                &                      &                 & $28.5$          \\
    Mallya~\textit{et al.}~\cite{transfer16}            & \checkmark                                &                      &                 & $33.8$          \\
    Girdhar~\textit{et al.}~\cite{attentionalpooling17} &                                           & \checkmark           &                 & $34.6$          \\
    Pairwise-Part \cite{pairwise18}                          & \checkmark                                & \checkmark           &                 & $39.9$          \\
    PastaNet \cite{pastanet20}                          & \checkmark                                & \checkmark           & \checkmark      & $46.3$          \\
    HAKE \cite{hake19}                          & \checkmark                                & \checkmark           & \checkmark      & $47.1$          \\
    \midrule
    DEFR (ours)                                         &                                           &                      &                 & $\textbf{60.5}$ \\
    \bottomrule
  \end{tabular}
\end{table}

\begin{table}[t]
  \caption{\textbf{The path from baseline to DEFR} evaluated on HICO. The baseline uses ImageNet-1K pre-trained ViT-B/32 as backbone, a random initialized linear classifier and binary cross entropy loss. Text embeddings from BERT or CLIP's text encoder are used for embedding initialization.}
  \label{tab:my_label}
  \centering
  \begin{tabular}{llccc}
    \toprule
    \textbf{} & \textbf{pre-train} & \textbf{LSE-Sign Loss} & \textbf{Embedding Init.} & \textbf{mAP}  \\ \midrule
    Baseline  & CLS1K             &                          &                        & 37.8          \\ \midrule
              & CLS1K             & \checkmark               &                        & 44.1          \\
              & CLS1K             & \checkmark               & ~\checkmark~BERT      & 53.5          \\
              & CLS1K             & \checkmark               & \checkmark~CLIP      & 54.7          \\
    DEFR      & CLIP              & \checkmark               & \checkmark~CLIP      & \textbf{60.5} \\
    \bottomrule
  \end{tabular}
\end{table}

\subsection{Ablations}
Several ablations were performed, focusing on the key components of DEFR: (a) pre-training tasks for the backbone, (b) initialization of the linear classifier, and (c) the loss function.

\textbf{From baseline to our solution:} \autoref{tab:my_label} shows the path from baseline to our DEFR.
The baseline uses ImageNet pre-trained ViT-B/32 as backbone, random initialization for the classifier, and binary cross entropy loss. It achieves 37.8 mAP, much lower than the detection-based approach (see \autoref{tab:mAP}). The LSE-Sign loss gains 6.3 mAP, and embedding initialization adds another 10.6 mAP using CLIP's text encoder. Further, when the text encoder in embedding initialization are jointly trained with the backbone (e.g. CLIP), the performance further improves by 5.8 mAP. It worth mentioning that fine-tuning the CLIP pre-trained backbone with a randomly initialized classifier scores only 36.8 mAP, much lower than the final solution. This demonstrates that CLIP pre-training, embedding initialization and LSE-Sign loss are effective and complementary mechanisms for detection-free HOI recognition.

\textbf{pre-training:}
We evaluate three different pre-training tasks on the backbone: (a) image classification (CLS1K/CLS21K), (b) masked language modeling (MLM), and (c) image-text contrastive learning (CLIP). \autoref{tab:pre-train_init} (the first column) shows the results for these pre-training tasks using random initialization in the linear classifier.

\textbf{Classifier Initialization:}
\autoref{tab:pre-train_init} compares random and embedding initialized linear classifier for four different pre-trained backbones. As described in \autoref{sec:embedding_init}, the embedding initialization is generated by CLIP or BERT text encoders from the class label text. Clearly, the embedding-based initialization provides a consistent improvement over all pre-training methods. It is worth noting that ImageNet pre-trained models (CLS1K/CLS21K) gain 10+ points from embedding initialization.  We believe that the text encoders effectively models the relationship between HOI classes such that embedding initialization makes feature learning easier in all pre-trained backbones.

\begin{table}[t]
  \centering
  \caption{\textbf{Ablations on backbone pre-training and classifier initialization}. Numbers are mAP on the HICO dataset. \textit{Random}: initialize the linear classifier with random weights; \textit{Embedding}: embedding initialization using BERT or CLIP's text encoder to generate embeddings. The backbone is ViT-B/32 and fine-tuned with LSE-Sign loss. Note that with CLIP text embedding initialization, all pre-trained backbones perform equally or better than the detection-supervised state of the art (47.1 mAP for HAKE \cite{hake19})}
  \label{tab:pre-train_init}
  \begin{tabular}{l c c c}
    \toprule
    \multirow{2}{*}{Pre-training} & \multicolumn{3}{c}{Classifier Initialization}                        \\
    \cmidrule(lr){2-4}
                                 & Random            & BERT Embedding             & CLIP Embedding      \\
    \midrule
    CLS1K                        & $44.1$            & $53.5_{(+9.4)}$            & $54.7_{(+10.6)}$      \\
    CLS21K                       & $44.2$            & $53.9_{(+9.7)}$            & $55.1_{(+10.9)}$      \\
    MLM                          & $43.6$            & $47.0_{(+3.4)}$            & $47.1_{(+3.5)}~~$      \\
    CLIP                         & $36.8$            & ~$51.0_{(+14.2)}$      & $\bm{60.5}_{(+23.7)}$ \\
    \bottomrule
  \end{tabular}
\end{table}

\begin{table}[t]
  \centering
  \caption{\textbf{Binary Cross Entropy (BCE) Loss vs. LSE-Sign Loss} evaluated on HICO for four different pre-trained models. The classifier is initialized with CLIP text embeddings.}
  \label{tab:bce_vs_sign}
  \begin{tabular}{l c c}
    \toprule
    \multirow{2}{*}{Pre-training} & \multicolumn{2}{c}{Loss Function}                        \\
    \cmidrule(lr){2-3}
                                 & BCE Loss       & LSE-Sign Loss      \\
    \midrule
    CLS1K                        & $51.5$       & $54.7_{(+3.2)}$      \\
    CLS21K                       & $50.0$       & $55.1_{(+5.1)}$      \\
    MLM                          & $46.6$       & $47.1_{(+0.5)}$      \\
    CLIP                         & $57.9$       & $\bm{60.5}_{(+2.6)}$ \\
    \bottomrule
  \end{tabular}
\end{table}

\textbf{Loss Function:}
The choice of loss function is vital in fine-tuning. Previous works use binary cross entropy (BCE) loss and treats HOI recognition as a set of binary classification problems. \autoref{tab:bce_vs_sign} shows that the proposed LSE-Sign loss outperforms BCE loss on all four pre-training backbones. \autoref{tab:clip_loss} compares LSE-Sign loss with other alternatives: binary cross entropy (BCE) loss, weighted BCE loss and focal loss~\cite{lin2017focal} on our best model.
Weighted cross entropy loss is intended to impose a weight on each class so that each category is balanced among the positive samples and negative samples.
Focal loss~\cite{lin2017focal} reduces the weight of the massive negative samples. Our LSE-Sign loss outperforms all others by a clear margin.

\textbf{Scalar $\gamma$ in \autoref{eq:si}:} LSE-Sign loss has a scalar $\gamma$ which controls the magnitude of output per class $s_i$. \autoref{tab:loss_gamma} shows the accuracy under different values of $\gamma$. The maximum performance is achieved when $\gamma=100$.

\begin{table}[t]
  \caption{\textbf{Comparison between LSE-Sign loss and other loss functions} evaluated on HICO. We change the loss function of the best model which has CLIP pre-trained ViT-B/32 backbone and embedding initialization. Weighted BCE is the binary cross entropy loss weighted by positive-negative-ratio per-class. Focal loss uses $\gamma$=2 and $\alpha$=0.25 as recommended in \cite{lin2017focal}.}
  \label{tab:clip_loss}
  \centering
  \begin{tabular}{lc}
    \toprule
    {\textbf{Loss function}} & {\textbf{mAP}}      \\
    \midrule
    Weighted BCE           & $54.7$        \\
    BCE                    & $57.9$        \\
    Focal Loss             & $53.2$        \\
    \midrule
    LSE-Sign Loss (ours)   & \textbf{60.5} \\
    \bottomrule
  \end{tabular}
\end{table}

\begin{table}[t]
  \caption{\textbf{Ablation of scalar} $\gamma$ in \autoref{eq:si} evaluated on HICO dataset. The highest accuracy is achieved at $\gamma=100$. The backbone is CLIP pre-trained ViT-B/32, classifier is embedding initialized and LSE-Sign loss is used for fine-tuning.}
  \label{tab:loss_gamma}
  \centering
  \begin{tabular}{lccccc}
    \toprule
    $\bm{\gamma}$ & 50   & \textbf{100} & 150  & 300  & 500  \\
    \midrule
    \textbf{mAP}      & 60.4 & \textbf{60.5} & 59.1 & 57.2 & 53.3 \\
    \bottomrule
  \end{tabular}
\end{table}

\subsection{Discussion of Limitations}

Although this paper reveals that image-text pre-training (e.g. CLIP) and LSE-Sign loss function play important roles for detection-free HOI recognition, we have not shown if these two components can work together with the detection supervision to further push the state-of-the-arts. Another limitation is that our method does not perform equally well for classes with very limited training samples.
We will investigate these in the future work.

\section{Conclusion}
In this paper, we presented a detection-free solution (DEFR) for human-object interaction (HOI) recognition. It not only simplifies the pipeline significantly, but also achieves higher accuracy than the detection-supervised counterparts. DEFR builds upon two findings. Firstly, we show that classifier initialization is crucial. Using the text embeddings of class labels generated by CLIP (image-text pre-trained) text encoder boosts the performance even for backbones pre-trained on images alone. Secondly, we propose a LSE-Sign loss to facilitate learning from multiple labels. Our detection-free method achieves a significant improvement (13.4 mAP) over the detection-supervised state of the art. We hope that this method opens up a new direction for HOI recognition.

\section*{Broader Impact}
In this paper, we shift the paradigm of Human-Object Interaction recognition from \textit{detection-supervised} to \textit{detection-free}, which substantially simplifies the training/inference pipeline and saves the computational resources spent on object and human keypoint detection. This could be a new direction for HOI recognition. Our work has great potential in many vision applications that require human activity understanding such as surveillance, retail, and health care systems. Furthermore, since our method is simple and requires less computations, it is more practical to integrate it into the real-time systems.

{\small\bibliography{refs}}
\end{document}